\documentclass{article}

\usepackage{PRIMEarxiv}

\usepackage[utf8]{inputenc} 
\usepackage[T1]{fontenc}    
\usepackage{hyperref}       
\usepackage{url}            
\usepackage{booktabs}       
\usepackage{amsfonts}       
\usepackage{nicefrac}       
\usepackage{microtype}      
\usepackage{lipsum}
\usepackage{fancyhdr}       
\usepackage{graphicx}       
\graphicspath{{media/}}     

\title{Building Height Prediction with Instance Segmentation
}

{\author{
  Furkan Burak BAĞCI, Ahmet Alp KINDIROĞLU, Metehan YALÇIN, Ufuk UYAN, Mahiye Uluyağmur ÖZTÜRK \\
  Huawei Turkey R\&D Center \\
  Istanbul, Turkey\\
  \texttt{\ Corresponding Author Email: furkan.burak.bagci@huawei.com} 
  \And
 \\
}}

\begin{document}
\maketitle

\begin{abstract}
Extracting building heights from satellite images is an active research area used in many fields such as telecommunications, city planning, etc. Many studies utilize DSM (Digital Surface Models) generated with lidars or stereo images for this purpose. Predicting the height of the buildings using only RGB images is challenging due to the insufficient amount of data, low data quality, variations of building types, different angles of light and shadow, etc. In this study, we present an instance segmentation-based building height extraction method to predict building masks with their respective heights from a single RGB satellite image. We used satellite images with building height annotations of certain cities along with an open-source satellite dataset with the transfer learning approach. We reached, the bounding box mAP 59, the mask mAP 52.6, and the average accuracy value of 70\% for buildings belonging to each height class in our test set.
\end{abstract}


\section{Introduction}

\subsection{Problem Definition}
\label{sec:headings}

Landcover mapping is an important subject used in many fields such as telecommunications, city planning, and transportation. This problem used to be costly, as people had to traverse the target area with measuring instruments. In addition, since there may be changes in some objects over time (for example, many buildings are demolished and new ones are built over time), the mapping process becomes out of date after a certain period of time. With the development of remote sensing technologies and deep learning-based computer vision methods, object detection and image segmentation applications on satellite images have been more accurate. Therefore, the landcover mapping problem can be solved fast and cheaply.

One particular field of landcover mapping is the calculation of height data from remote images. Due to the lack of publicly available height information for buildings, there exists only a handful of studies on problems such as building height prediction. Most of the height prediction studies use stereo images or DSM that are important for depth extraction. It is valuable to extract height information from monocular images where we do not have such other data sources. This is a challenging task as depth information depends on angle of satellite and light, shadows that can change based on time of day and occlusions of larger objects on smaller objects etc. In this  study our hypothesis is given sufficient large building height image dataset, the model can learn to predict building height despite these problems.

\subsection{Instance Segmentation}
\label{sec:headings}
Many studies have been carried out on the extraction of buildings from satellite images with semantic segmentation methods. Semantic segmentation tools can predict object masks at the pixel level, however, it is not possible to distinguish between objects in the same class. Some post-processing operations (thresholding, connected component analysis, etc.) are required so that each building can be individually identified and modeled in 3D. However, the connections between the semantic masks of neighboring buildings may cause us to treat them as a single building and this may result in incorrect 3D modeling.

Instance segmentation is a challenging computer vision problem that aims to predict the bounding box and mask for each object in the input image at the instance level. The instance segmentation methods consist of components from object detection and semantic segmentation methods. It provides the separation of each object in the same class.

YOLOACT \cite{bolya2019yolact}, a real-time, fully convolutional instance segmentation method, runs at 33.5 fps on Titan Xp and reaches 33.5 mAP on the MS COCO dataset. This method treats the instance segmentation problem as two main subtasks. These are, generating set of prototype masks and estimating the coefficients for each object mask. Final mask estimates are determined by linear combination of prototype masks with coefficients. Mask2Former \cite{cheng2021mask2former}, a transformer-based approach, has achieved new state-of-the-art results on the MS-COCO dataset in panoptic, instance and semantic segmentation tasks. The main component in this method is masked attention, which extracts local features from predicted masks by constraining cross-attention

Mask R-CNN \cite{he2017mask}, which is based on a two-stage Faster R-CNN object detection method, is proposed as a general framework and can be used for instance segmentation and pose estimation. It is a powerful method and it is easy to adapt it to different domains. The main difference between Mask R-CNN from Faster R-CNN is the segmentation branch as an addition to classification and bounding box regression branches, which can work in parallel on each region proposal.

\subsection{Literature Review}
\label{sec:headings}

In our literature review, we did not find any study using instance segmentation for building and height prediction. However, some studies try to detect buildings without height prediction.

Chitturi et. al.\cite{chitturi2020building} proposed a single-class building instance segmentation model from satellite images with Mask RCNN. In their study, some TTA(Test Time Augmentation) techniques (horizontally-vertically flip, bight, contrast) were applied and the effect of each augmentation on model performance was reported. Fritz \cite{fritz2020instance} compared various semantic (Unet and FCN) and instance segmentation (Mask R-CNN) architectures for building prediction. In \cite{li2021novel}, authors proposed a novel segmentation framework based on Mask R-CNN and histogram thresholding for classifying new and old buildings. Zhou et. al. \cite{zhou2019building} trained Mask R-CNN and FCN (Fully Convolutional Network) networks using very high resolution aerial images and compared the results or building segmentation tasks. As a result, they reported that Mask R-CNN is better in detecting mask edges and works with 15\% higher accuracy than FCN in detecting small objects. Since many studies suggest Mask R-CNN-based methods for building instance segmentation problem, we based this method in our study.

In another group of studies, authors try to predict heights from pixel level height annotation with DSM. 
The Karatsiolis \cite{karatsiolis2021img2ndsm} and  Liu \cite{liu2020im2elevation} both proposed a UNET-based custom building height segmentation model. They tried to estimate the height at the pixel level from the RGB image in a regression-like way with the loss of RMSE and MAE.

\section{Method}
\label{sec:headings}

\subsection{Preprocessing}
\label{sec:headings}
There are many problems in the raw data set, such as overlapping annotations, incorrect or missing annotations, and annotation shifts. To overcome these problems, we performed some data preprocessing operations.

\subsubsection{Overlapping Anotations}
\label{sec:headings}
This problem occurs by overlapping small size high length annotations on high peak areas of the building and low length annotations on the rest of the building for a high height building such as a skyscraper and plaza etc. In this case, training our network with the same part of a building as both high and low length, decreases the model accuracy. For this reason, we developed an algorithm to merge multiple overlapping annotations with the tallest one.

The IOU (Intersection Over Union) is one of the most widely used metrics for calculating how well model predictions fit with ground truth masks in segmentation and bounding boxes in object detection. The IOU is calculated by the ratio of the overlapping area of two object predictions to the total area. However, IOU does not work well to decide if two annotations overlap because this value is too small for annotations of very different sizes and the algorithm fails for most of the overlapping buildings. For this reason, we used intersection over each annotation area, to decide whether two annotations overlap. The metric we use, calculates the ratio of intersection over each annotation area and uses the maximum one. In this way, when a large part of any mask intersects with the other small mask, those annotations are merged.

\subsection{Misaligned and Partial Annotations}
\label{sec:headings}
Another problem in the dataset is incorrect annotations and shifted ones. Considering that there are tens of thousands of images in the data set, it is not a realistic solution to manually filter the faulty samples. Therefore, erroneous samples in our data set were filtered by an automatic filtering method. In our experiments, it has been shown that the Unet++ semantic segmentation model works well against such faulty examples. For this reason, first of all, a single-class Unet++ building segmentation model is trained. Then, we made inferences with this model on our whole data set. After that, we calculated the IOU between each of the predicted buildings and each of the annotations to determine missing predictions and annotations. If the ratio of missing predictions or annotations is above 50\% then we discarded that image.

\subsection{Over-Under Detections}
\label{sec:headings}
In our data set, some small buildings are annotated as a group. If these buildings are detected one by one, they are considered to be incorrect. Similarly, for some groups of buildings, our method could predict a single building. Calculating only one-to-one matches causes us to ignore some of the acceptable model predictions for the business use case of the model. For this reason, over-detection and under-detection cases need to be considered for evaluation.

\begin{figure}[!htb]
    \centering
    \includegraphics[scale=0.8]{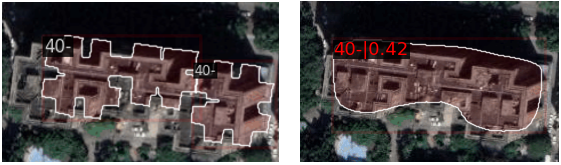}
    \caption{Under Detection Case}
   \label{over-under det}
\end{figure}

Over-detection refers to situations when there are multiple predictions for an object, while under-detection occurs when there is one prediction for multiple objects as in Figure \ref{over-under det}. Ozdemir et. al.\cite{ozdemir2010performance} proposed an approach that calculates over-detection and under-detection in object detection performance measurement. We used the approach from this work while calculating the confusion matrix. Firstly, we identified 1-to-1 matches. Then we calculated over-detection and under-detection cases. In the under-detection calculation, for a 1-to-1 mismatch, we detected all annotations in the same class with an IOU value greater than 0. Then, the new object formed by combining these annotations is calculated. Lastly, the IOU value between this combined object and our prediction was calculated. If the IOU value is higher than the threshold value, that prediction is also considered correct. A similar scenario has been applied for over-detection cases as well.

\subsection{Training}
\label{sec:headings}
In the training phase, we used the open-source mmdetection \cite{mmdetection} library, which includes Pytorch implementation of many object detection and instance segmentation methods. In order to apply the transfer learning approach, we first train a single class building instance segmentation model with an open-source building segmentation dataset \cite{7yfp-9p87-22} which was proposed by Miyazaki. Then we fine-tuned those weights with our 3 class dataset, by initializing our model with them.

We used the Resnet-50 backbone network for our initial experiments because of its speed. After performing the hyperparameter optimization, we ran our experiments with the ResNext-101 backbone, which outperformed Resnet-50.

\subsubsection{Transfer Learning}
\label{sec:headings}
Since different geographies contain different types of objects, model generalization is a problem.

Our dataset, consists of satellite images from certain regions along with buildings and related heights. The instance segmentation model, trained using only our dataset, performs well only in a limited region, but generalizes poorly to other regions.

In order to overcome this issue, it is important to use the features of the buildings from images taken in different light conditions and different building types from very wide geographies. Recently published data set \cite{7yfp-9p87-22} has been used for transfer training because of its good features such as; the quality of the building labels, the separation of the labels of different buildings from each other, and the geographical variation of images.

The images and annotations in the open-source dataset were first divided into 512x512 parts. Then original building annotations, generated for semantic segmentation were converted for instance segmentation using adaptive thresholding and connected component analysis.

In the first step, the Mask R-CNN model was trained with an open source single class building data set. The MS-COCO weights were used as a starting point for the encoder. In the pretraining phase, only the first layer of the ResNext encoder is frozen and all other layers are trained. After the learning rate and other hyperparameters are optimized, we fine-tuned this model with our 3 class-building height dataset by initializing the weights obtained in the pre-training as a starting point. In the fine-tuning part, the encoder is frozen, and the other layers of the network are trained.

For both pretraining and fine tuning, we increased the number of anchor boxes and added smaller anchor boxes because the ratio of building annotations to the image is very small in the satellite dataset compared to the objects in the MS-COCO dataset. We changed the default scales parameter of the Region proposal network from [8] to [2, 4, 8, 16, 32].

\section{Results}
\label{sec:headings}
In this section, we first describe the datasets that we used. Afterthat, we provided experimental results.

\subsection{Data Set}
In our experiments, we make use of two different datasets to train our model. The first one, which we pretrain our model, is the public dataset proposed by Miyazaki. The second dataset, we used for fine-tuning contains building masks and respective heights for certain regions.
\label{sec:headings}

\subsubsection{Public Data Set}
\label{sec:headings}
The public aerial dataset \cite{7yfp-9p87-22} has 18127 images with 1024x1024 resolution from 7 different countries and pixel-level building annotations. We used images from 4 countries (Japan, Thailand, Kenya, and Mozambique) from this dataset. We ran our experiments with about 20000 images generated as 512x512 slices from original full-size images.

\subsubsection{Height Data Set}
\label{sec:headings}
Our private data set consists of 744 satellite images with 5000x6000 resolution, segmentation masks for each building, and height values for each building belonging to 5 different cities. In order to reduce the computational complexity and increase the training speed and accuracy, we prepared 512x512 grid-shaped slices and performed training and testing on these images. Our dataset has 3 classes and these are; up to 15 meters, between 15 meters and 40 meters, and over 40 meters. We selected our labels after make histogram analysis on our dataset, considering the balance between classes. As we were looking for a solution to the building height prediction problem with instance segmentation, we converted our raw input data into coco annotation format, which is one of the well-known formats.

\subsection{Experiments}
\label{sec:headings}
We calculated the performance of our algorithm on randomly selected 1200 test images from a city. We used the mAP value along with the confusion matrix for the evaluation.


\begin{figure}[!htb]
   \begin{minipage}{0.5\textwidth}
     \centering
     \includegraphics[width=0.8\linewidth]{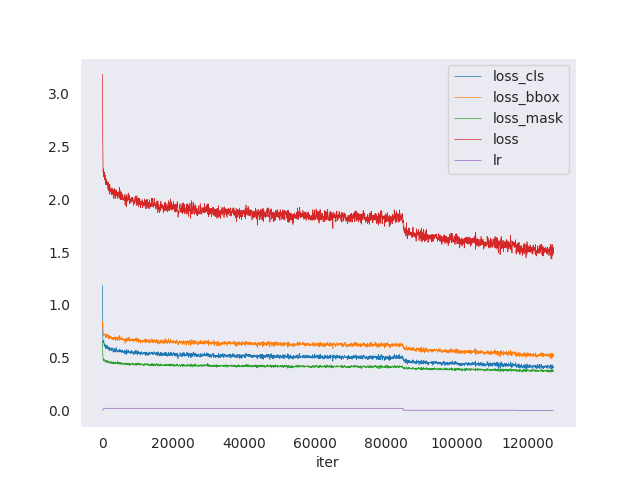}
     \caption{Train Loss}\label{Fig:Data1}
   \end{minipage}\hfill
   \begin{minipage}{0.5\textwidth}
     \centering
     \includegraphics[width=0.8\linewidth]{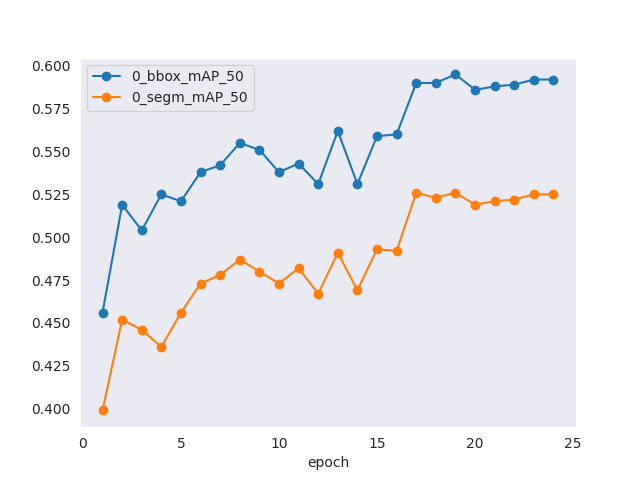}
     \caption{Validation mAP}\label{Fig:Data2}
   \end{minipage}
\end{figure}

The bounding box mAP(IOU=0.5) value of the model was 59 and the segmentation mask mAP(IOU=0.5) value was 52.6 as shown in Figure \ref{tab:confusion_matrix}.

\begin{table}[!htb]
 \caption{Confusion Matrix}
  \centering
  \begin{tabular}{lllll}
    \toprule
    \multicolumn{5}{c}{Prediction}                   \\
    \cmidrule(r){2-5}
    Ground Truth     & 0m-15m     & 15m-40m  & 40m+ & Background\\
    \midrule
    0m-15m      & \textbf{73\%}  & 8\%  & 0\%  & 18\% \\
    15m-40m    & 16\%   & \textbf{71\%}  & 5\%  & 6\% \\
    40m+         & 1\%   & 23\%  & \textbf{65\%}  & 9\% \\
    Background & 88\%   & 9\%  & 2\%   & 0\% \\
    \bottomrule
  \end{tabular}
  \label{tab:confusion_matrix}
\end{table}

The model reached 73\% accuracy for buildings up to 15 meters, 71\% for buildings between 15 meters and 40 meters, and 65\% for buildings over 40 meters. Most of the mistakes belong to adjacent classes. For example, 8\% of buildings that belong to up to 15 meters class were estimated between 15 meters and 40 meters, but none were estimated above 40 meters.
False positive predictions were most common in buildings up to 15 meters class that model confuses at most.

\section{Conclusion}
\label{sec:headings}

Building height prediction is a demanding problem in many sector including telecommunication, transprotation etc. Predicting height from single view RGB images is a challenging task, because of the light reflection, effect of the big shadows, lack of view from differnet perspectives for low height buildings etc. We proposed instance segmentation-based building height prediction solution from single RGB images. Most of the studies utilize semantic segmentation architectures to predict pixel-based buildings, however, our model not only tries to predict buildings but also their heights at the instance level. We applied transfer learning and fine-tuning approach in order to overcome the model generalization problem. Our model reached 59 bbox mAP and 52.6 segmentation mAP values and in overall it reached about 70\% accuracy for all of the height classes.


\end{document}